\documentclass{article} 
\usepackage{arxiv2026,times}

\usepackage{amsmath,amsfonts,bm}

\def\eqref#1{equation~\ref{#1}}

\def\1{\bm{1}}

\DeclareMathAlphabet{\mathsfit}{\encodingdefault}{\sfdefault}{m}{sl}
\SetMathAlphabet{\mathsfit}{bold}{\encodingdefault}{\sfdefault}{bx}{n}

\usepackage{hyperref}
\usepackage{url}

\usepackage{tcolorbox}
\usepackage{tcolorbox}
\tcbuselibrary{skins,breakable}
\usepackage{booktabs}  
\usepackage{graphicx} 
\usepackage{amssymb}   
\usepackage{bbding}    
\usepackage{algorithm}
\usepackage{algpseudocode}
\usepackage{amsmath}
\usepackage{color}
\usepackage{multirow}
\usepackage{booktabs}
\usepackage{multirow}
\usepackage{xcolor}
\usepackage{colortbl}
\usepackage{wrapfig}

\usepackage{graphicx}
\algdef{SE}[DOWHILE]{Do}{doWhile}{\algorithmicdo}[1]{\algorithmicwhile\ #1}
\algtext*{Do}
\algtext*{doWhile}

\title{KnowGuard: Knowledge-Driven Abstention for Multi-Round Clinical Reasoning}

\author{Xilin Dang\textsuperscript{1}\thanks{Equal contribution}, Kexin Chen\textsuperscript{1,2*}, Xiaorui Su\textsuperscript{2}, Ayush Noori\textsuperscript{2}, Iñaki Arango\textsuperscript{2}, Lucas Vittor\textsuperscript{2},  \\ 
\textbf{Xinyi Long\textsuperscript{3}, Yuyang Du\textsuperscript{3}, Marinka Zitnik\textsuperscript{2}\thanks{Marinka Zitnik is the corresponding author.}, Pheng Ann Heng\textsuperscript{1}}  \\
\textsuperscript{1}Department of Computer Science and Engineering, The Chinese University of Hong Kong\\
\textsuperscript{2}Department of Biomedical Informatics, Harvard Medical School\\
\textsuperscript{3}Department of Information Engineering, The Chinese University of Hong Kong\\
\texttt{marinka@hms.harvard.edu}
}

\arxivfinalcopy
\begin{document}

\maketitle

\begin{abstract}

In clinical practice, physicians refrain from making decisions when patient information is insufficient. This behavior, known as abstention, is a critical safety mechanism preventing potentially harmful misdiagnoses. Recent investigations have reported the application of large language models (LLMs) in medical scenarios. However, existing LLMs struggle with the abstentions, frequently providing overconfident responses despite incomplete information. This limitation stems from conventional abstention methods relying solely on model self-assessments, which lack systematic strategies to identify knowledge boundaries with external medical evidences. To address this, we propose \textbf{KnowGuard}, a novel \textit{investigate-before-abstain} paradigm that integrates systematic knowledge graph exploration for clinical decision-making. Our approach consists of two key stages operating on a shared contextualized evidence pool: 1) an evidence discovery stage that systematically explores the medical knowledge space through graph expansion and direct retrieval, and 2) an evidence evaluation stage that ranks evidence using multiple factors to adapt exploration based on patient context and conversation history. This two-stage approach enables systematic knowledge graph exploration, allowing models to trace structured reasoning paths and recognize insufficient medical evidence. We evaluate our abstention approach using open-ended multi-round clinical benchmarks that mimic realistic diagnostic scenarios, assessing abstention quality through accuracy-efficiency trade-offs beyond existing closed-form evaluations. Experimental evidences clearly demonstrate that KnowGuard outperforms state-of-the-art abstention approaches, improving diagnostic accuracy by 3.93\% while reducing unnecessary interaction by 7.27 turns on average.

\end{abstract}

\section{Introduction}

Large language models (LLMs) are designed to generate prompt responses based on given instructions~\citep{brown2020language}. However, in clinical decision-making, this tendency becomes problematic, as patient's initial information is often incomplete or ambiguous, requiring iterative, multi-round conversations to be progressively disclosed. In such scenarios, the ability to abstain, i.e., recognizing knowledge boundaries and refraining from answering under uncertainty, is crucial for ensuring the safe and effective deployment of clinical AI systems. Yet, current LLMs struggle with abstention, frequently providing overconfident or premature responses. This behavior prolongs diagnostic interactions, delays decision-making, and increases the cognitive burden on physicians, ultimately undermining trust in AI-assisted workflows~\citep{sun2025large,kumaran2025overconfidence}.

\begin{figure}[ht]
\centering
\includegraphics[width=1\textwidth]{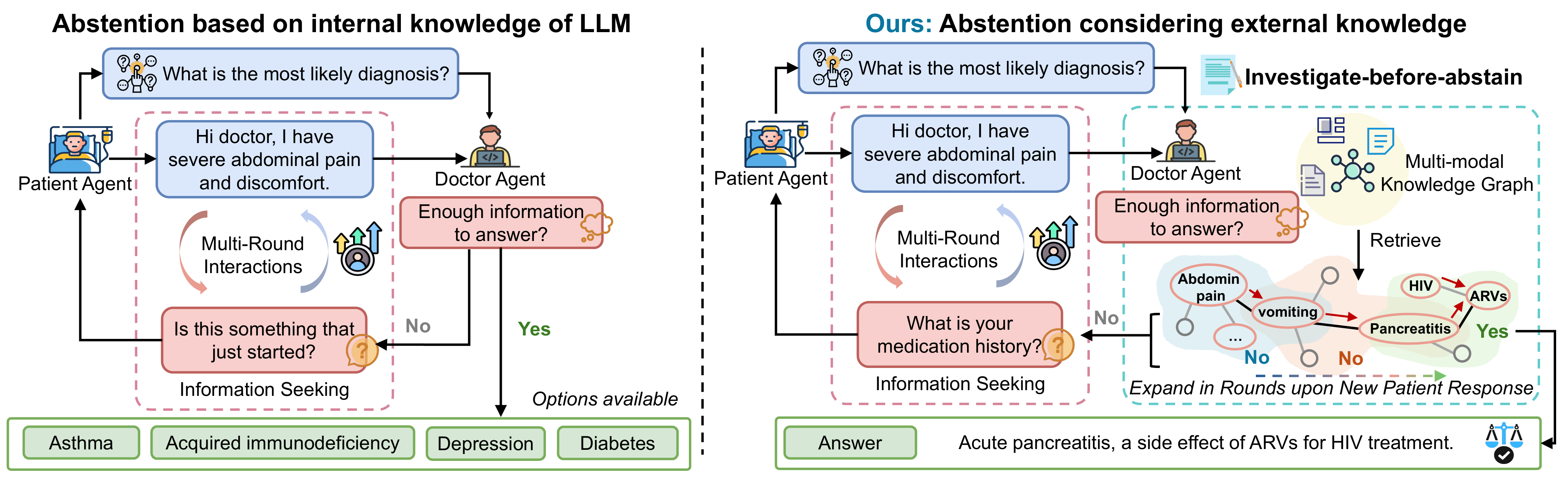}
\caption{Comparison of abstention approaches in multi-round clinical reasoning. Traditional methods (left) rely on confidence assessment using internal LLM knowledge. Our \textit{investigate-before-abstain} paradigm (right) proactively detects knowledge boundaries through systematic medical knowledge graph exploration, identifying evidence gaps to guide targeted investigation before abstention decisions.}
\label{fig:teaser}  
\end{figure}

Existing abstention methods face two fundamental challenges that limit their suitability for clinical applications. First, LLMs inherently exhibit overconfidence and choice-supportive bias. Traditional confidence-based methods~\citep{tian2023just,li2024mediq,geng2023survey} rely on LLM self-assessments to generate confidence scores for abstentions. However, LLMs often inflate their confidence in initial answers, even when faced with contradictory evidence~\citep{tian2025overconfidence}. This issue could be further exacerbated by the model's reasoning fine-tuning, a post-training method that has been widely applied in recent medical agents~\citep{kirichenko2025abstentionbench}. This overconfidence becomes particularly problematic in multi-round clinical conversations, where models maintain false certainty despite limited patient information. Second, current methods lack robust external knowledge validation methods. Even evidence collection methods, such as the one reported in ~\citep{srinivasan2024selective}, count on internal model knowledge without referencing external medical knowledge. These limitations prove especially concerning in clinical settings, where life-critical decisions require both higher reliability and systematic reasoning grounded in external, verifiable evidence.

\textbf{Present work: }This paper incorporates external medical knowledge to address the abstention problem, aiming to ground the LLM's abstention decisions with factual medical evidence beyond its own understanding. The key implementation challenge of the proposed approach is to efficiently and precisely identify the knowledge boundary, i.e., determining whether available evidence is sufficient to support a reliable conclusion. In light of this, a highly structured data representation of the external knowledge source is required to facilitate easier and more accurate boundary identifications. Knowledge graph provides well-organized medical relationships, and is, therefore, a good match to support the systematic reasoning needed for our abstention approach ~\citep{gao2025leveraging,pan2024unifying}.

We highlight that the abstention problem requires a systematic exploration of the medical knowledge graph beyond simple fact retrieval. Under a practical multi-round setup, the system must maintain investigation consistency across interactions and dynamically adapt to new patient information provided. To this end, we propose a novel \textit{investigate-before-abstain} paradigm that grounds abstention decisions in systematic exploration of medical knowledge graphs. This approach progressively investigates knowledge boundaries across rounds, integrating external knowledge with clinical abstention. When new patient details emerge, the system continues exploration rather than restarting, using knowledge conflicts as signals of uncertainty (see Figure~\ref{fig:teaser} for details). Our approach consists of two major stages operating on a shared contextualized evidence pool. The \textit{evidence discovery stage} queries and updates knowledge triplets through graph expansion and direct retrieval based on new patient information. The \textit{evidence evaluation stage} ranks evidence using multiple factors including graph coherence, embedding similarity, LLM selection, temporal decay, and patient population reasoning to identify reliable evidence and facilitate contextualized abstention assessment. Throughout multi-round interactions, this evidence pool functions as a priority queue, continuously updating evidence relevance based on evolving patient context.
In summary, this paper puts forth \textbf{KnowGuard}, a multi-round clinical question answering (QA) abstention approach that leverages knowledge graphs with contextualized evidence reasoning. Our major contributions are summarized as follows: 
(1) \textbf{\textit{Investigate-before-abstain} paradigm}: We replace the unreliable LLM self-assessment scheme with our systematic medical knowledge graph exploration, grounding abstention decisions in factual evidence.
(2) \textbf{Multi-round knowledge graph reasoning}: We design a two-stage approach with evidence discovery through graph expansion and direct retrieval, followed by evidence evaluation using coherence-aware scoring and demographic-guided reasoning that enables dynamic knowledge expansion adapted to evolving patient information. 
(3) \textbf{Dataset and benchmark}: We establish a new open-ended multi-round clinical benchmark comprising 3,061 cases across three medical datasets. Additionally, we construct a comprehensive medical knowledge graph derived from over 300 WHO guidelines. This knowledge graph encompasses 22k nodes and over 100k edges, integrating multimodal information across text, image, and relation. Unlike existing clinical QA datasets that use multiple-choice formats, our open-ended setting better reflects real clinical conversations and enables proper evaluation of abstention behavior.
(4) \textbf{Comprehensive system evaluation}: We compare against 5 representative abstention baselines with and without enhancement techniques. Extensive comparisons with state-of-the-art abstention approach show that our method improves diagnostic accuracy by 3.93\% while reducing interaction rounds by 7.27 turns on average.

\section{Related Work}

\noindent\textbf{Medical Question Answering Systems.}
LLM-powered agents have advanced medical question answering (QA)~\citep{jin2021disease,singhal2023large,su2024kgarevion}, which encompasses both multiple-choice and open-ended questions from diverse medical sources. 
To better reflect real-world clinical practice where physicians often need to gather additional information through iterative questioning, recent research has shifted toward interactive QA frameworks that allow for multi-turn conversations and information seeking~\citep{wang2025healthq,johri2025evaluation,li2024mediq}.
MediQ~\citep{li2024mediq} introduced such an interactive QA framework that leverages multi-agent collaboration to encourage agents to abstain from answering when uncertain and actively seek additional information through follow-up questions.
However, existing interactive benchmarks predominantly focus on multiple-choice formats, which inadequately reflect real-world clinical scenarios where practitioners typically encounter open-ended questions without predefined answer choices~\citep{nachane2024few}. 
To address this limitation, we develop a multi-round open-ended interactive clinical reasoning benchmark to evaluate free-text responses.

\noindent\textbf{Abstention Methods.} 
Effective abstention requires recognizing knowledge boundaries and refraining from answering when evidence is insufficient~\citep{lin2025explore,ni2025towards,kale2025line}. Current approaches include self-assessment methods that rely on internal confidence through uncertainty estimation~\citep{tian2023just}, calibration scoring~\citep{geng2023survey,srivastava2023beyond}, and multi-scale rating~\citep{li2024mediq}; consistency-based methods that aggregate multiple model outputs for disagreement detection~\citep{wang2022self}; and knowledge-based approaches that incorporate information sources. Long context methods~\citep{tu2024towards} retrieve comprehensive medical documents but provide coarse-grained context that fails to pinpoint specific knowledge gaps, leading to information overload rather than targeted evidence discovery.
While these methods have shown promise in various domains, they share a fundamental limitation in their reliance on \textit{reactive confidence assessment} rather than \textit{proactive knowledge investigation}. When facing uncertainty, these methods ask ``how confident am I?'' instead of ``what specific evidence am I missing?''.
KnowGuard introduces the first \textit{investigate-before-abstain} paradigm for multi-round clinical reasoning, which systematically explores knowledge boundaries through targeted evidence discovery guided by medical knowledge graphs.

\section{Method}

\begin{figure}[h]
\centering
\includegraphics[width=1\textwidth]{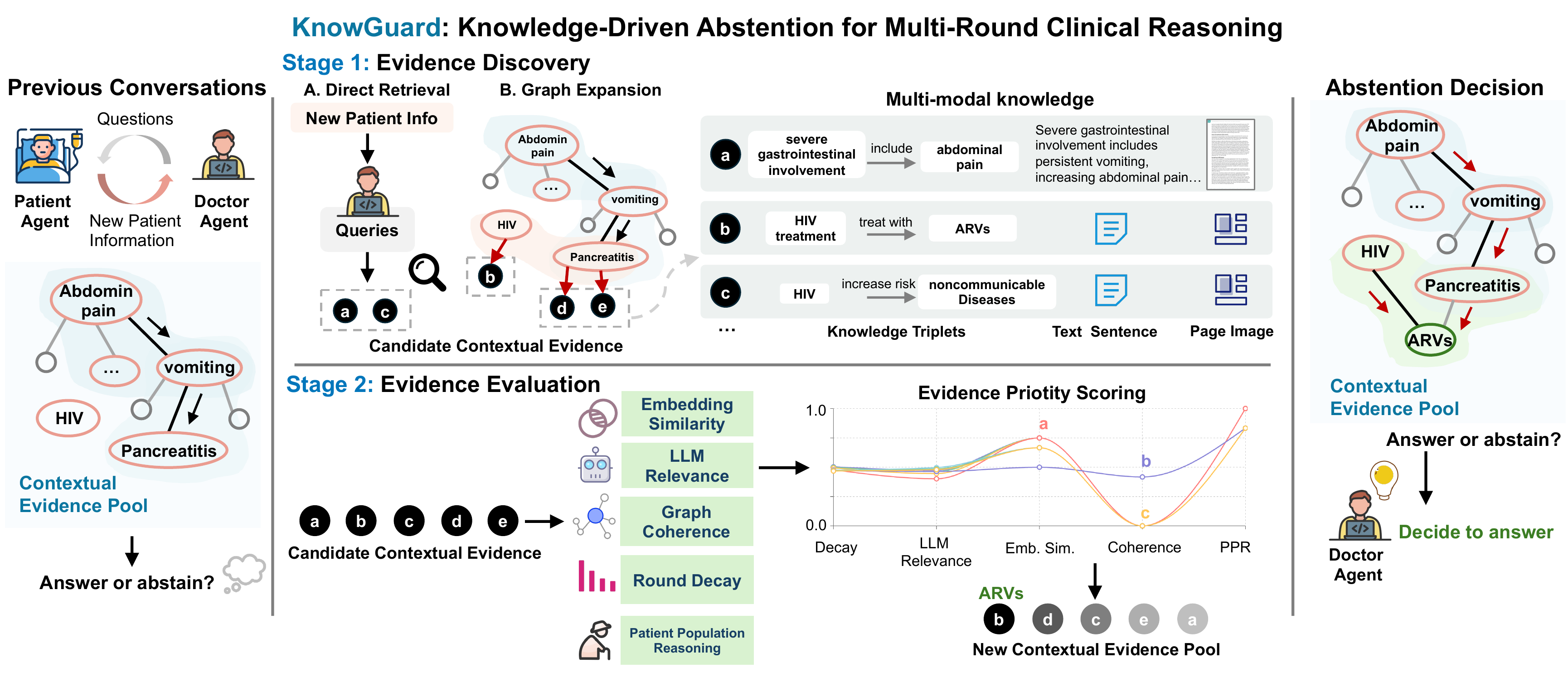}
\caption{
KnowGuard approach for knowledge-driven abstention in clinical reasoning. Our \textit{investigate-before-abstain} paradigm systematically explores medical knowledge graphs to identify evidence gaps before abstention decisions. The Evidence Discovery Stage retrieves multi-modal evidence through dynamic graph expansion and direct retrieval. The Evidence Evaluation Stage adapts exploration priorities through relevance assessment, graph coherence prioritization, demographic weighting, and temporal decay. Final abstention decisions integrate all factors to determine when sufficient evidence exists for diagnosis versus continued investigation.
}
\label{fig:overview}
\end{figure}

\subsection{Problem Formulation and approach Overview}

\noindent\textbf{Multi-round Abstention Problem Formalization.} 
We formalize multi-round clinical abstention within an interactive consultation approach that simulates realistic diagnostic scenarios. The Patient Agent maintains complete patient information $\mathcal{K} = \{k_0, k_1, \ldots, k_n\}$ ($n$ pieces in total) and responds truthfully to inquiries by revealing relevant information subsets. 
The Doctor Agent receives the initial patient presentation $k_0$ and must decide at each round $t$ whether to abstain from diagnosis. When abstaining, the agent asks targeted questions $q_t$ to gather additional information; otherwise, it provides a diagnostic answer.
At each round $t$, given accumulated patient knowledge $\mathcal{K}_t = \mathcal{K}_{t-1} \cup \{a_t\}$ where $a_t$ represents the patient's response to question $q_t$, the Doctor Agent must make a binary abstention decision:
\begin{equation}
\mathcal{A}_t: \mathcal{K}_t \rightarrow \{0, 1\},
\end{equation}
where $\mathcal{A}_t = 0$ indicates continued information gathering (abstention) and $\mathcal{A}_t = 1$ indicates sufficient confidence for diagnosis. 
The core challenge lies in determining the optimal stopping point where $\mathcal{K}_t$ contains sufficient evidence for reliable diagnosis while minimizing unnecessary interaction rounds. Our proposed method focuses on this challenge.

\noindent\textbf{KnowGuard Approach.} 
Our \textit{investigate-before-abstain} paradigm replaces unreliable LLM self-assessment with structured medical knowledge exploration. As shown in Figure~\ref{fig:overview}, KnowGuard maintains a contextualized evidence pool $\mathcal{B}_t$ represented as a priority queue of knowledge triplets relevant to the case. 
The evidence pool evolves cumulatively across conversation rounds, building upon previous discoveries while incorporating new patient information $a_t$.
The approach operates through two complementary stages: Evidence Discovery Stage systematically expands $\mathcal{B}_t$ based on patient information, while Evidence Evaluation Stage adapts exploration priorities based on multiple factors, including patient demographics.

\noindent\textbf{Multi-modal Knowledge Graph Construction.} 
We construct a comprehensive medical knowledge graph $\mathcal{G} = (\mathcal{V}, \mathcal{E})$ from authoritative medical guidelines, containing $|\mathcal{V}|$ medical entities and $|\mathcal{E}|$ clinical relationships. Each triplet $(h, r, t) \in \mathcal{E}$ is augmented with source text descriptions and document page images, enabling both structured reasoning and contextual validation during evidence discovery.

\subsection{Evidence Discovery Stage}

\noindent\textbf{Contextualized Evidence Pool Definition.} 
The Evidence Discovery Stage operationalizes knowledge boundary exploration through systematic graph investigation.
To enable efficient exploration of vast medical knowledge spaces, we maintain a contextualized evidence pool as a priority queue $\mathcal{B}_t = \{(h_i, r_i, t_i, p_i)\}_{i=1}^{|\mathcal{B}_t|}$ of candidate medical triplets (length is $K$), where each triplet $(h_i, r_i, t_i)$ represents a potential reasoning step with priority $p_i$. This bounded representation enables efficient ranking and selection while focusing exploration on the most promising knowledge paths.

\noindent\textbf{Systematic Evidence Expansion.}
The stage performs structured exploration through two complementary retrieval strategies. Graph Expansion-based retrieval identifies triplets connected to entities in current high-priority candidates:
\begin{equation}
\mathcal{T}_{\text{exp}} = \{(h, r, t) \in \mathcal{G} : h \in \mathcal{E}_{\mathcal{B}_t} \text{ or } t \in \mathcal{E}_{\mathcal{B}_t}\},
\end{equation}
where $\mathcal{E}_{\mathcal{B}_t}$ represents entities present in current evidence triplet. Direct retrieval first generates queries according to the current patient response $a_t$, and then performs a comprehensive search across the knowledge graph:
\begin{equation}
\mathcal{T}_{\text{query}} = \text{GraphRetrieval}(\mathcal{G}, \text{LLM}_{\text{query}}(a_t)).
\end{equation}

The retrieved evidence candidates $\mathcal{T}_{\text{candidates}} = \mathcal{T}_{\text{exp}} \cup \mathcal{T}_{\text{query}}$ are fed into the evidence evaluation stage for priority scoring.

\subsection{Evidence Evaluation Stage}
The Evidence Evaluation Stage operates on candidate contextual evidence to compute comprehensive priority scores through five complementary factors: Embedding similarity, LLM relevance, graph coherence, round decay, and patient population reasoning.

\noindent\textbf{Relevance Assessment with Dual Validation.} 
Each candidate triplet undergoes dual relevance assessment combining embedding similarity (hard relevance) and LLM relevance (soft relevance). Hard relevance measures the semantic similarity between triplet embeddings and the current patient response:
\begin{equation}
s_{\text{sim}}(h, r, t) = \text{cosine}(\text{Embed}(h, r, t), \text{Embed}(a_t)),
\end{equation}
while soft relevance employs LLM to assess clinical relevance given the current patient context:
\begin{equation}
s_{\text{rel}}(h, r, t) = \text{LLM}_{\text{rel}}(a_t, (h, r, t)).
\end{equation}
This dual validation ensures both semantic and clinical alignment of evidence investigation.

\noindent\textbf{Graph Coherence Prioritization.} 
To maintain reasoning consistency, we prioritize triplets that connect to frequently visited entities, indicating established reasoning pathways:
\begin{equation}
s_{\text{coh}}(h, r, t) = \text{count}_{\mathcal{B}}(h) + \text{count}_{\mathcal{B}}(t),
\end{equation}
where $\text{count}_{\mathcal{B}}(\cdot)$ tracks cumulative frequency of the entity across all evidence pools throughout the conversation. Higher coherence scores indicate stronger integration with existing paths, enabling systematic knowledge boundary detection rather than random exploration.

\noindent\textbf{Demographic-guided Priority Weighting.} 
The stage infers patient demographics and clinical populations from conversation history to prioritize relevant knowledge graph regions. Population inference analyzes accumulated patient information against predefined categories:
\begin{equation}
\mathcal{P}_t = \text{LLM}_{\text{demo}}(\mathcal{K}_t, \mathcal{C}_{\text{pop}}),
\end{equation}
where $\mathcal{C}_{\text{pop}}$ represents predefined population categories derived from knowledge graph topics, such as adolescents.
Triplets belonging to identified patient populations receive enhanced weighting:
\begin{equation}
s_{\text{pop}}(h, r, t) = \begin{cases}
\alpha & \text{if } (h, r, t) \in \text{Subgraph}(\mathcal{P}_t) \\
1 & \text{otherwise},
\end{cases}
\end{equation}
where $\alpha > 1$ emphasizes population-specific knowledge and $\text{Subgraph}(\mathcal{P}_t)$ denotes triplets relevant to inferred populations.

\noindent\textbf{Temporal Decay with Round-based Updates.} 
To balance historical context with current information, the stage applies temporal decay to previously explored knowledge while emphasizing recent evidence. Priority updates follow exponential decay:
\begin{equation}
p_{t+1}(h, r, t) = p_t(h, r, t) \times (1 - w_{\text{decay}}) + p_{\text{new}}(h, r, t) \times w_{\text{decay}},
\end{equation}
where $p_{\text{new}}$ reflects priority computed from current round information and $w_{\text{decay}} \in [0, 1]$ controls temporal transition rate.

\noindent\textbf{Evidence-grounded Abstention Decision.}
The final priority combines multiple contextual factors through weighted aggregation:
\begin{equation}
p_{\text{final}}(h, r, t) = (w_{\text{sim}} \cdot s_{\text{sim}} + w_{\text{rel}} \cdot s_{\text{rel}} + w_{\text{coh}} \cdot s_{\text{coh}}) \times s_{\text{pop}}.
\end{equation}
The contextualized evidence pool maintains top-$K$ triplets: $\mathcal{B}_{t+1} = \text{Top-K}(\mathcal{T}_{\text{candidates}}, p_{\text{final}})$, where $\mathcal{T}_{\text{candidates}} = \mathcal{T}_{\text{exp}} \cup \mathcal{T}_{\text{query}}$. Each triplet is augmented with multi-modal evidence including source text and document images. The final abstention decision integrates structured knowledge evidence with patient context:
\begin{equation}
    \mathcal{A}_t = \text{LLM}_\text{doctor}(\mathcal{K}_t, \mathcal{B}_t, \{x_{\text{text}}, x_{\text{img}}\}),
\end{equation}
where the model receives current patient information, top-ranked evidence triplets, and their associated multi-modal content to make informed abstention decisions.

\subsection{Open-Ended Clinical Reasoning Benchmark}

To properly evaluate abstention behavior in realistic clinical scenarios, we establish a multi-round open-ended benchmark that extends beyond existing closed-form evaluations. Traditional multiple-choice formats constrain response options and fail to capture the complexity of real clinical conversations where physicians must formulate comprehensive diagnostic assessments. Following recent advances in automated evaluation~\citep{su2024kgarevion}, we employ LLM-as-judge methodology to convert closed-ended questions to an open-ended format, enabling more accurate assessment of both diagnostic reasoning quality and abstention appropriateness.

The Judge Agent performs answer matching between free-text predictions and ground truth responses. For originally multiple-choice questions, the judge receives all answer options along with the model's free-text response, without knowing the question content or correct option, and identifies the most semantically similar option:
\begin{equation}
\mathcal{A}_{\text{matched}} = \text{Judge}(\mathcal{A}_{\text{pred}}, \{\text{option}_1, \text{option}_2, \ldots, \text{option}_n\}).
\end{equation}
For originally open-ended questions, the judge performs binary classification to determine whether the prediction aligns with the ground truth answer:
\begin{equation}
\text{Match} = \text{Judge}(\mathcal{A}_{\text{pred}}, \mathcal{A}_{\text{true}}) \in \{\text{Yes}, \text{No}\},
\end{equation}
where $\mathcal{A}_{\text{pred}}$ represents the model's free-text response and $\mathcal{A}_{\text{true}}$ denotes the ground truth answer.

\section{Experiments and Results}
We conducted extensive experiments to evaluate the effectiveness of KnowGuard on multi-round clinical abstention, comparing against existing abstention methods on our open-ended interactive clinical reasoning benchmark.

\subsection{Experimental Settings}

\noindent\textbf{Dataset Construction.}
We convert MEDQA (CC-BY-4.0)~\citep{jin2021disease}, CRAFT-MD (CC-BY-4.0)~\citep{johri2024craft}, and AFRIMEDQA (CC-BY-NC-SA-4.0)~\citep{nimo2025afrimed} into interactive multi-round formats. 
Following established protocols~\citep{li2024mediq}, we parse patient records into structured components: age, gender, chief complaint, and additional evidence as atomic facts~\citep{min2023factscore}. Initially, only age, gender, and chief complaint are presented to the Doctor Agent, which must strategically gather missing information through targeted questioning. The resulting interactive datasets are termed ioMEDQA, ioCRAFT-MD, and ioAFRIMEDQA.

\noindent\textbf{Multi-modal Knowledge Graph Construction.}
Our knowledge graph incorporates over 300 WHO guidelines, resulting in 22k medical entities and more than 100k clinical relationships. Each triplet is augmented with source text and document images for comprehensive knowledge boundary detection. Subgraphs are labeled with demographic and disease-specific features extracted from guideline titles and abstracts, enabling patient population reasoning. The system monitors publication dates for automatic updates, ensuring current medical knowledge supports boundary detection decisions.

\noindent\textbf{Baseline Methods.}
We benchmark KnowGuard against representative abstention approaches: Basic (direct question or answer, without explicit abstention step), Binary Decision~\citep{srivastava2023beyond} (explicit binary abstention), Numerical Score~\citep{tian2023just} (confidence scoring 1-5 with thresholding), Scale Rating~\citep{li2024mediq} (fine-grained confidence levels with descriptions), and Long Context~\citep{tu2024towards} (external document retrieval with full-text processing).
We compare the baselines with and without rationale generation~\citep{wei2022chain} (generate rationale alongside abstention decision) and self-consistency~\citep{wang2022self} as enhancements.

\noindent\textbf{Metrics and Agent.} We evaluate using Accuracy (ACC) and average conversation rounds (avg. Turn) as primary metrics for diagnostic effectiveness and interaction efficiency. All experiments employ GPT-4~\citep{achiam2023gpt} as the core agent model, given its widespread adoption and demonstrated capabilities in medical reasoning tasks~\citep{eriksen2024use}.

\subsection{Results}

\begin{table}
\centering
\scalebox{0.68}{
\begin{tabular}{lccccccc}
\toprule
\multirow{2}{*}{Method} & \multicolumn{2}{c}{ioAFRIMEDQA} & \multicolumn{2}{c}{ioMEDQA} & \multicolumn{2}{c}{ioCRAFT-MD} \\
 \cmidrule(lr){2-3} \cmidrule(lr){4-5} \cmidrule(lr){6-7}
& ACC & avg. Turn & ACC & avg. Turn & ACC & avg. Turn \\
\midrule
\multicolumn{7}{c}{\textbf{Basic Methods Comparison}} \\
\midrule
Basic (implicit) & $51.10\pm 2.40$ & $8.32\pm 0.43$ & $57.83\pm 2.05$ & $8.98\pm 0.32$ & $54.69\pm 1.27$ & $8.31\pm 0.26$ \\
Binary Decision~\citep{srivastava2023beyond} & $61.97\pm 2.83$ & $8.98\pm 0.54$ & $65.95\pm 1.87$ & $7.69\pm 0.33$ & $64.67\pm 1.13$ & $7.85\pm 0.30$ \\
Numerical Score~\citep{tian2023just} & $54.25\pm 2.69$ & $1.72\pm 0.27$ & $61.74\pm 1.76$ & $2.51\pm 0.17$ & $59.35\pm 1.20$ & $2.42\pm 0.26$ \\
Scale Rating~\citep{li2024mediq} & $63.06\pm 2.34$ & $5.11\pm 0.48$ & $64.23\pm 1.53$ & $5.15\pm 0.23$ & $65.40\pm 1.19$ & $4.83\pm 0.21$ \\
Long Context~\citep{tu2024towards} & $57.45\pm 2.08$ & $2.01\pm 0.18$ & $59.95\pm 1.20$ & $3.23\pm 0.22$ & $57.88\pm 1.58$ & $3.23\pm 0.14$ \\
KnowGuard & $\mathbf{68.70\pm 1.77}$ & $5.26\pm 0.61$ & $\mathbf{70.98\pm 1.98}$ & $5.41\pm 0.15$ & $\mathbf{66.47\pm 1.47}$ & $4.89\pm 0.17$ \\
\midrule
\multicolumn{7}{c}{\textbf{Enhanced Methods with Rationale Generation~\citep{wei2022chain} and Self-Consistency~\citep{wang2022self}}} \\
\midrule
Binary Decision~\citep{srivastava2023beyond} & $64.55\pm 2.99$ & $13.82\pm 0.56$ & $72.92\pm 1.47$ & $13.00\pm 0.42$ & $70.01\pm 1.35$ & $12.21\pm 0.33$ \\
Numerical Score~\citep{tian2023just} & $58.33\pm 2.79$ & $2.63\pm 0.45$ & $64.23\pm 1.72$ & $4.61\pm 0.30$ & $61.51\pm 1.17$ & $4.98\pm 0.35$ \\
Scale Rating~\citep{li2024mediq} & $61.36\pm 1.00$ & $5.31\pm 0.05$ & $65.52\pm 1.36$ & $6.26\pm 1.13$ & $66.34\pm 1.89$ & $5.56\pm 0.17$ \\
Long Context~\citep{wang2024beyond} & $56.80\pm 0.33$ & $1.16\pm 0.48$ & $59.37\pm 0.84$ & $3.30\pm 1.15$ & $58.61\pm 0.85$ & $3.29\pm 0.97$ \\
KnowGuard & $\mathbf{73.20\pm 1.92}$ & $5.30\pm 0.58$ & $\mathbf{74.12\pm 0.57}$ & $5.40\pm 0.27$ & $\mathbf{71.96 \pm 0.98}$ & $6.51\pm 0.09$ \\
\bottomrule
\end{tabular}
}
\caption{Performance comparison on open-ended multi-round interactive clinical reasoning. Accuracy and average turns are reported for baseline methods and their enhanced versions.}
\label{tab:open-acc-comparison}
\end{table}

Table~\ref{tab:open-acc-comparison} demonstrates KnowGuard's superior performance across all benchmarks. Our \textit{investigate-before-abstain} paradigm achieves the highest accuracy while maintaining competitive interaction efficiency, systematically identifying knowledge gaps rather than relying on self-assessments. KnowGuard consistently outperforms all baseline methods, achieving 1.07-5.64\% accuracy improvements over the strongest confidence-based approaches (Binary Decision and Scale Rating) in basic settings, and 1.20-8.65\% improvements in enhanced settings. Compared to knowledge-enhanced Long Context, KnowGuard delivers substantial gains of 10.29\% accuracy in basic settings and 14.83\% in enhanced settings on average.
Notably, while Long Context also incorporates external knowledge, it retrieves comprehensive documents without systematic boundary detection, leading to information overload and premature abstention decisions. The integration of rationale generation and self-consistency benefits all methods, with KnowGuard showing 3-4\% accuracy improvements while maintaining stable interaction lengths, demonstrating the robustness of knowledge boundary detection over self-assessment-based abstention approaches.

\section{Analysis}

\subsection{Ablation Studies on Key Components}

\begin{table}
    \centering
    \caption{
    Ablation studies of KnowGuard's key designs, including evidence modality of text or multi-modal knowledge graph (KG) triplet, evidence evaluation stage (Evidence Eval.), and patient population reasoning (PPR) factor.
    }
    \scalebox{0.8}{
    \begin{tabular}{cccccccccc}
        \toprule
        \multicolumn{4}{c}{Component Configuration} & \multicolumn{2}{c}{ioAFRIMEDQA} &
        \multicolumn{2}{c}{ioMEDQA} &
        \multicolumn{2}{c}{ioCRAFT-MD} \\
          Text evidence & KG evidence & Evidence Eval. & PPR & ACC & avg. Turn & ACC &
          avg. Turn & ACC & avg. Turn \\
        \midrule 
        \Checkmark & \Checkmark & \Checkmark & \Checkmark &
         73.20 & 5.30 & 74.12 & 5.40 & 71.96 & 6.51 \\
        \cmidrule(lr){1-10}
        \Checkmark & \Checkmark & \Checkmark & \XSolidBrush &
        72.60 & 7.03 & 74.29 & 6.53 & 71.92 & 6.53 \\
        \Checkmark & \Checkmark & \XSolidBrush & \XSolidBrush &
        66.22 & 2.69 & 70.66 & 3.24 & 68.92 & 3.31 \\
        \Checkmark & \XSolidBrush& \XSolidBrush&\XSolidBrush &
        66.02 & 3.33 & 64.79 & 3.25 & 62.73 & 3.30 \\
        \XSolidBrush & \XSolidBrush & \XSolidBrush & \XSolidBrush & 63.06 & 5.11 & 64.23 & 5.15 & 65.40 & 4.83 \\
        \bottomrule
    \end{tabular}}
    \label{tab:ablation}
\end{table}

To validate the effectiveness of KnowGuard's designs, we conducted systematic ablation studies as shown in Table~\ref{tab:ablation}. We progressively evaluate each component's contribution to demonstrate their individual effectiveness.
Multi-modal knowledge graph triplets provide substantial improvements over text-only evidence retrieval, demonstrating the value of structured medical knowledge for abstention. The evidence evaluation stage enables systematic exploration by ranking candidate evidence, leading to more targeted abstention decisions. Patient Population Reasoning (PPR) enhances personalized reasoning by considering demographic and disease-specific contexts. Each component contributes meaningfully to both accuracy and efficiency, with the complete system achieving optimal performance across all datasets.

\subsection{Hyperparameter Studies}

\begin{table}[h]
\centering
\caption{Sensitivity analysis of evidence evaluation factors. Embedding similarity is abbreviated as Embed. Sim.}
\label{tab:hyperparameter_sensitivity}
\scalebox{0.88}{%  % 缩放到80%
\begin{tabular}{l|c|cc|cc|cc}
\toprule
\multirow{2}{*}{\textbf{Factor Weight}} & \multirow{2}{*}{\textbf{Value}} & \multicolumn{2}{c|}{\textbf{ioAFRIMEDQA}} & \multicolumn{2}{c|}{\textbf{ioCRAFT-MD}} & \multicolumn{2}{c}{\textbf{ioMEDQA}} \\
\cmidrule{3-8}
& & ACC & Round & ACC & Round & ACC & Round\\
\midrule
\multirow{3}{*}{Embed. Sim.} 
& 0.10 & 71.41 &5.48 & 71.10 & 5.44 & 72.77 & 5.14 \\
& 0.20 & 73.20 &5.30 & 71.96 & 5.51 & 74.12 & 5.40 \\
& 0.30 & 71.99 &5.46 & 71.67 & 5.40 & 71.24 & 5.15 \\
\midrule
\multirow{3}{*}{LLM Relevance} 
& 0.50 & 71.02 & 5.58 & 69.58 & 5.44 & 70.59 & 5.25 \\
& 0.60 & 73.20 & 5.30 & 71.96 & 5.51 & 74.12 & 5.40 \\
& 0.70 & 68.51 & 5.40 & 70.22 & 5.37 & 71.64 & 5.23 \\
\midrule
\multirow{3}{*}{Graph Coherence} 
& 0.25 & 71.41 & 5.35 & 69.50 & 5.45 & 70.27 & 5.17 \\
& 0.35 & 73.20 & 5.30 & 71.96 & 5.51 & 74.12 & 5.40 \\
& 0.45 & 70.44 & 5.65 & 72.47 & 5.43 & 71.72 & 5.17 \\
\midrule
\multirow{3}{*}{Round Decay} 
& 0.40 & 70.25 & 5.41 & 68.78 & 5.51 & 70.84 & 5.33 \\
& 0.50 & 73.20 & 5.30 & 71.96 & 5.51 & 74.12 & 5.40 \\
& 0.60 & 68.32 & 5.33 & 71.51 & 5.29 & 71.16 & 5.18 \\
\midrule
\multirow{3}{*}{PPR} 
& 1.10 & 70.25 & 5.55 & 70.30 & 5.47 & 71.64 & 5.20 \\
& 1.15 & 73.20 & 5.30 & 71.96 & 5.51 & 74.12 & 5.40 \\
& 1.20 & 71.22 & 5.38 & 70.24 & 5.39 & 71.98 & 5.23 \\
\bottomrule
\end{tabular}%
}
\end{table}

Our evidence priority scoring mechanism combines multiple factors for systematic exploration. Table~\ref{tab:hyperparameter_sensitivity} shows sensitivity analysis for each factor. All factors contribute meaningfully to performance, with optimal weights being: embedding similarity $w_{\text{sim}}$ (0.2), LLM relevance $w_{\text{rel}}$ (0.6), graph coherence $w_{\text{coh}}$ (0.35), round decay $w_{\text{decay}}$ (0.5), and patient population reasoning $w_{\text{pop}}$ (1.15). The consistent performance across different weight configurations demonstrates the robustness of our approach, indicating that the method is not overly sensitive to hyperparameter tuning.

\begin{figure}[h]
\centering
\includegraphics[width=1\textwidth]{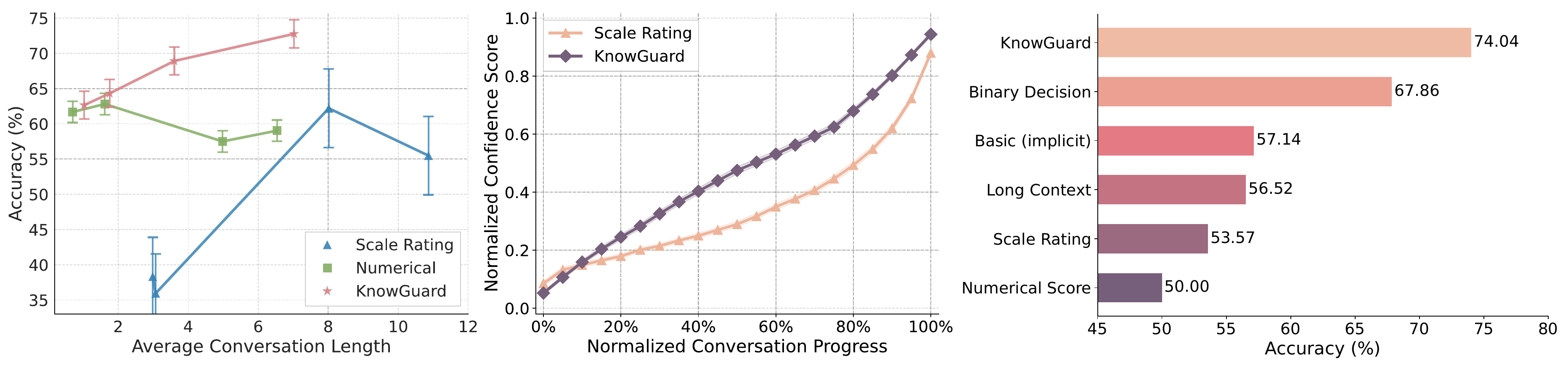}
\caption{(Left) Systematic evidence exploration enables accuracy improvements with longer conversations, unlike confidence-based self-assessment methods. (Middle) KnowGuard's confidence evolves more rapidly through targeted evidence acquisition compared to generic self-assessment. (Right) \textit{Investigate-before-abstain} paradigm particularly benefits rare disease diagnosis where external knowledge exploration is crucial.}

\label{fig:analysis}
\end{figure}

\noindent\textbf{Accuracy vs. Conversation Length.}
Figure~\ref{fig:analysis}(Left) demonstrates the relationships between accuracy and conversation length for our method and traditional self-assessment approaches (Scale Rating, Numerical Score). KnowGuard shows consistent accuracy improvements with longer conversations, indicating effective knowledge boundary investigation through systematic external knowledge exploration. In contrast, self-assessment methods show steep trajectories where additional rounds provide diminishing returns, reflecting their reliance on internal knowledge. This validates our core hypothesis that proactive knowledge exploration outperforms reactive confidence assessment in multi-round clinical reasoning.

\noindent\textbf{Confidence Evolution during Conversation.}
Figure~\ref{fig:analysis}(Middle) shows confidence evolution patterns of our method and Scale Rating throughout conversations. The lengths of different conversations are normalized for intuitive presentation and comparison. Notably, KnowGuard's confidence increases more rapidly than Scale Rating. This indicates that systematic exploration of medical knowledge boundaries enables more targeted information gathering than generic self-assessment. 

\noindent\textbf{Performance on Rare Cases.}
Figure~\ref{fig:analysis}(Right) compares the accuracy performance on rare diseases. KnowGuard demonstrates substantial advantages over other abstention methods. This suggests that introducing external knowledge as contextual evidence effectively enhances reasoning in challenging cases where traditional self-assessment methods struggle, while the design of patient population reasoning enables targeted exploration of relevant medical subgraphs for more informed abstention decisions.

\begin{figure}[h]
\centering
\includegraphics[width=1\textwidth]{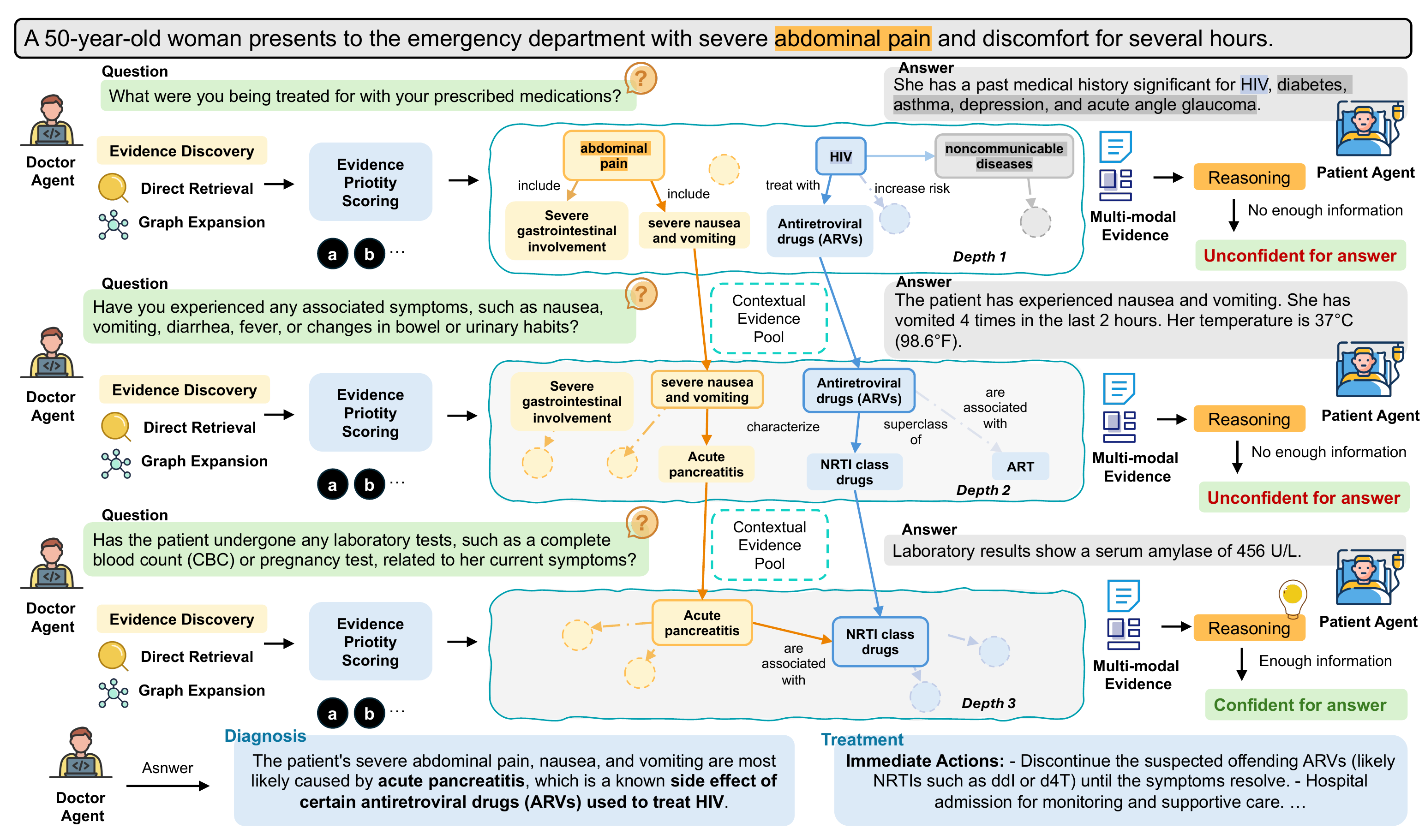}
\caption{Case study demonstrating KnowGuard's \textit{investigate-before-abstain} paradigm.}
\label{fig:case_study}
\end{figure}

\noindent\textbf{Case Study.}
Figure~\ref{fig:case_study} illustrates an example of KnowGuard's \textit{investigate-before-abstain} paradigm. When presented with abdominal pain symptoms, the system proactively investigates contextual evidence to explore medical knowledge boundaries. Through structured exploration guided by medical knowledge graphs, the system discovers the connection between NRTI-class drugs (used for HIV) and acute pancreatitis, ultimately reaching an accurate diagnosis with comprehensive treatment recommendations. This demonstrates how systematic knowledge boundary exploration enables confident decision-making in complex clinical scenarios.

\section{Conclusion}

In this work, we establish the novel task of open-ended multi-round clinical reasoning and present KnowGuard, an \textit{investigate-before-abstain} paradigm that shifts from internal LLM knowledge to external evidence investigation. Rather than asking ``how confident am I?" when facing uncertainty, KnowGuard systematically investigates ``what specific evidence am I missing?" through structured exploration of medical knowledge graphs.
Our comprehensive experiments demonstrate that this paradigm shift yields substantial improvements across diverse clinical datasets, with KnowGuard achieving state-of-the-art performance compared to five baseline abstention methods across all benchmarks.

\section{Ethics Statement}
KnowGuard is a research prototype not intended for clinical deployment. The system utilizes publicly available datasets and may exhibit biases inherent in underlying models and knowledge sources. Comprehensive fairness evaluations represent important future work.

\bibliography{reference}
\bibliographystyle{arxiv2026}

\end{document}